\documentclass{vgtc}

\ifpdf%
  \pdfoutput=1\relax                   %
  \usepackage{graphicx}                %
  \DeclareGraphicsExtensions{.pdf,.png,.jpg,.jpeg} %
\else%
  \usepackage{graphicx}                %
  \DeclareGraphicsExtensions{.eps}     %
\fi%

\graphicspath{{figures/}{pictures/}{images/}{./}} %

\usepackage{microtype}                 %
\PassOptionsToPackage{warn}{textcomp}  %
\usepackage{textcomp}                  %

\usepackage{hyperref}
\hypersetup{
    colorlinks, 
    linkcolor=blue
}

\usepackage{mathptmx}                  %
\usepackage{times}                     %
\usepackage{cite}                      %
\usepackage{tabu}                      %
\usepackage{booktabs}                  %
\usepackage{xspace}

\usepackage{enumitem}

\onlineid{0}

\vgtccategory{Research}

\vgtcinsertpkg

\title{\textsc{ManimML\xspace{}}: Communicating Machine Learning Architectures with Animation}

\author{Alec Helbling\thanks{e-mail: alechelbling@gatech.edu}\\ %
        \scriptsize Georgia Institute of Technology%
\and Duen Horng (Polo) Chau \thanks{e-mail: polo@gatech.edu}\\
\scriptsize Georgia Institute of Technology}

\teaser{
  \centering
  \includegraphics[width=\linewidth]{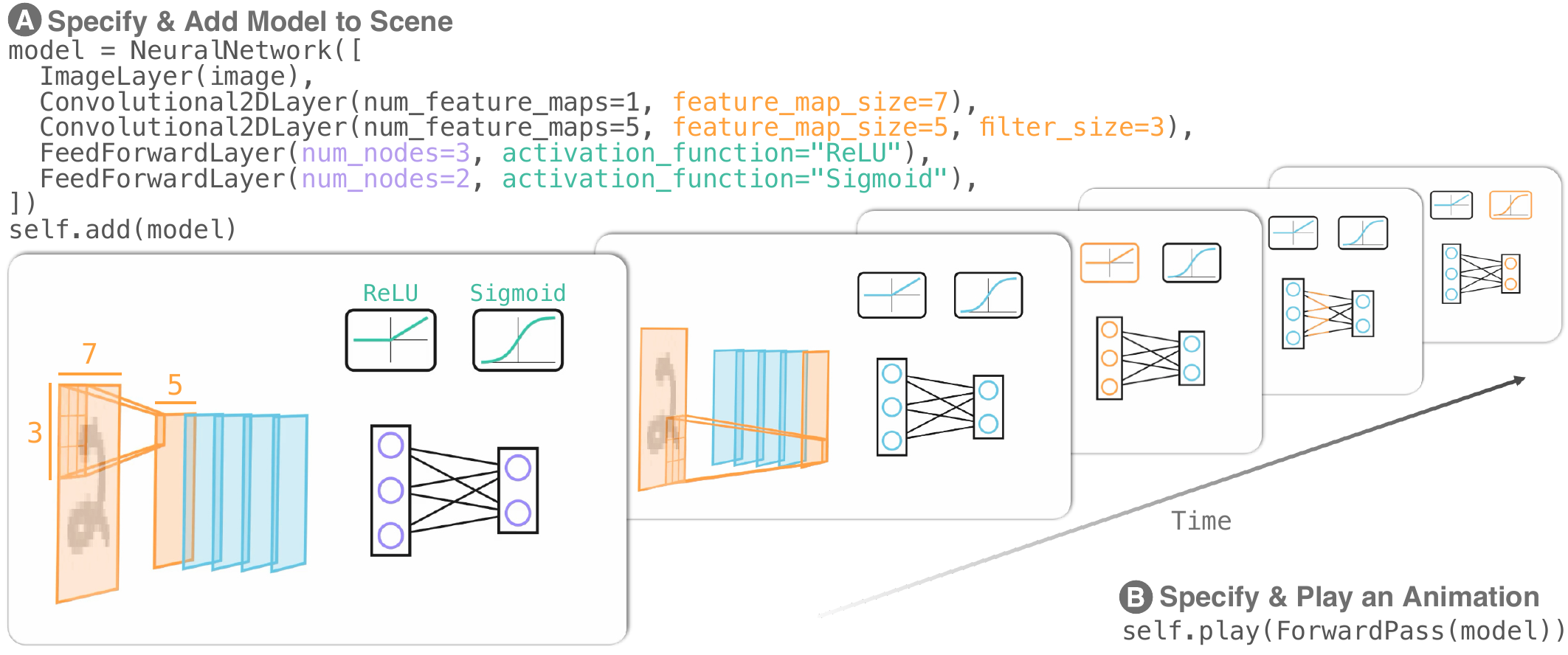}
  \vspace{-0.2in}
  \caption{\textsc{ManimML\xspace{}} rendering an animation of the forward pass of a convolutional neural network. \textbf{(A)} First a user specifies the neural network architecture with code using a familiar Pytorch-like syntax. Parameters are passed to each layer, specifying various information like the number of convolutional feature maps, the activation function, etc. The model is added to the scene and can be rendered as a single still frame (Bottom Left). \textbf{(B)} The user can then specify an animation they wish to play with code. Frames for the animation of the forward pass for the specified neural network are shown. An animation is automatically constructed by combining primitive animations for each pair of layers. First a convolution animation is shown, followed by a feed forward neural network animation.}
  \label{fig:teaser}
}

\abstract{
    There has been an explosion in interest in machine learning (ML) in recent years due to its applications to science and engineering. 
    However, as ML techniques have advanced, tools for explaining and visualizing novel ML algorithms have lagged behind. Animation has been shown to be a powerful tool for making engaging visualizations of systems that dynamically change over time, which makes it well suited to the task of communicating ML algorithms. However, the current approach to animating ML algorithms is to handcraft applications that highlight specific algorithms or use complex generalized animation software. We developed \textsc{ManimML\xspace{}}, an open-source Python library for easily generating animations of ML algorithms directly from code. We sought to leverage ML practitioners' preexisting knowledge of programming rather than requiring them to learn complex animation software.  \textsc{ManimML\xspace{}} has a familiar syntax for specifying neural networks that mimics popular deep learning frameworks like Pytorch. A user can take a preexisting neural network architecture and easily write a specification for an animation in \textsc{ManimML\xspace{}}, which will then automatically compose animations for different components of the system into a final animation of the entire neural network. \textsc{ManimML\xspace{}} is open source and available at \url{https://github.com/helblazer811/ManimML}. 

} %

\CCScatlist{
  \CCScatTwelve{Meachine Learning}{Visu\-al\-iza\-tion}{Deep Learning}{Visualization Systems and Tools};
}

\definecolor{orange}{RGB}{255,119,0}
\definecolor{red}{RGB}{220,0,0}
\definecolor{agreen}{RGB}{74, 198, 148}
\definecolor{purple}{RGB}{158, 62, 177}
\definecolor{darkpurple}{RGB}{170, 70, 210}
\definecolor{aqua}{RGB}{87, 180, 181}
\definecolor{lightblue}{RGB}{72, 123, 232}
\definecolor{hotpink}{RGB}{255, 83, 115}
\definecolor{teal}{RGB}{90, 200, 250}
\definecolor{linkColor}{RGB}{6,125,233}
\definecolor{RdColorscale}{RGB}{213, 96, 80}
\definecolor{BuColorscale}{RGB}{75, 148, 196}
\definecolor{BrColorscale}{RGB}{188, 132, 53}
\definecolor{BGColorscale}{RGB}{57, 152, 143}
\definecolor{LOrangesColorscale}{RGB}{255, 195, 133}
\definecolor{ROrangesColorscale}{RGB}{196, 65, 3}
\definecolor{layerGray}{RGB}{80, 80, 80}

\begin{document}

\firstsection{Introduction}

\maketitle

\begin{figure}
    \centering
    \includegraphics[width=\columnwidth]{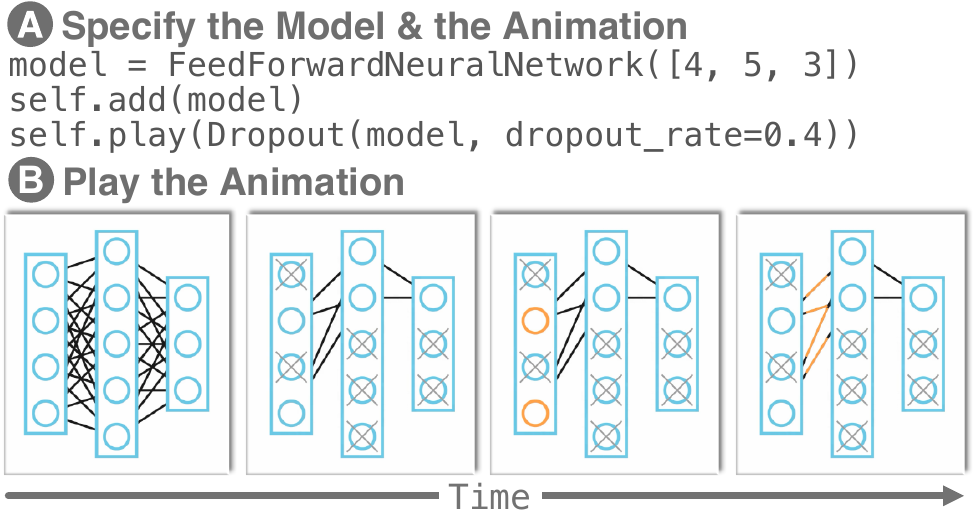}
    \vspace{-0.2in}
    \caption{\textsc{ManimML\xspace{}} is capable of animating dropout, a common neural network training technique. \textbf{(A)} Shows the minimal amount of code necessary to specify the neural network and desired animation. \textbf{(B)} Shows the animation that \textsc{ManimML\xspace{}} render's based on the user specification. }
    \label{fig:enter-label}
\end{figure}

Machine learning (ML) has garnered substantial interest in recent years due to its successful application to many areas of science. This explosion in interest in ML techniques has led to the development of tools for visualizing and explaining ML algorithms. A powerful technique for communicating the dynamic behavior of algorithms is animation, the efficacy of which has long been of interest to researchers \cite{brownTechniquesAlgorithmAnimation1985}. Animation has been shown to effectively increase learner engagement \cite{aminiHookedDataVideos2018, wangCNNExplainerLearning2021}, and can be especially helpful for showing transitions between the states of a system \cite{heerAnimatedTransitionsStatistical2007}. This makes it well suited to the task of communicating ML algorithms, which often involve many interacting components with complex sequential dependencies. Despite its studied efficacy, it remains difficult for ML practitioners to easily design faithful animations for explaining novel algorithms. Many existing ML visualization tools leverage animation \cite{wangCNNExplainerLearning2021, kahngGANLabUnderstanding2019}, however, they are often hand engineered to highlight the functionality of specific methodologies. This can be a prohibitively time-intensive process. 
Following these observations, our contributions are as follows:

\begin{enumerate}[topsep=5pt, itemsep=2mm, parsep=3pt, leftmargin=15pt]
    \item \textbf{\textsc{ManimML\xspace{}}, an open-source Python library for communicating ML architectures and the sequence of operations of ML algorithms with animation. } \textsc{ManimML\xspace{}} offers users the ability to quickly render animations of common ML architectures like neural networks using Python code. We chose Python because it is the default programming language for a plurality of ML researchers and practitioners \cite{cassTopProgrammingLanguages2021}, and is home to many popular deep learning libraries like Pytorch.\footnote{\url{https://pytorch.org/}} To make \textsc{ManimML\xspace{}} as familiar as possible to ML practitioners, our system for specifying neural networks to animate mimics the syntax of popular deep learning libraries (Figure \ref{fig:teaser}A). A user can specify a neural network architecture as an arbitrary sequence of neural network layers and \textsc{ManimML\xspace{}} will automatically construct a combined animation of the specified architecture that is faithful to the underlying algorithms (Figure \ref{fig:teaser}, \ref{fig:enter-label}). 
    \item \textbf{An extensible software library design that can be quickly adapted to new ML architectures. } Many existing ML visualization tools are hand-crafted to highlight information relevant to a specific algorithm \cite{wangCNNExplainerLearning2021, kahngGANLabUnderstanding2019}, which can be laborious and time-consuming. \textsc{ManimML\xspace{}} automatically composes a user specified collection of pre-implemented animations and visual assets, making it simple to generate animations of new systems, without needing to be built each piece from the ground up. Further, due to the rapid advancing nature of the ML literature it is important that our system supports the addition of new components and visualization techniques. The architecture of our library allows engineers to easily substitute in new animations and make extensions to existing components. 
\end{enumerate}

\vspace{-0.1in}
\section{System Design}
\textbf{Software Library Design.}
\textsc{ManimML\xspace{}} is an open-source software library released under the MIT license. We chose Python because of its popularity in the ML community. In addition to being more familiar to ML practitioners, Python is home to a wide array of ML and scientific computing software, which \textsc{ManimML\xspace{}} can potentially interface with. Our tool builds on top of the Manim Community (\url{https://www.manim.community/}) library, a Python tool for performing animations of mathematical concepts. Manim Community provides a set of abstractions and systems for creating vector graphics and other assets, placing them in a scene, and applying animations to them. \textsc{ManimML\xspace{}} is a major extension to Manim Community, offering the capability to animate ML algorithms and architectures. \textsc{ManimML\xspace{}} has a familiar syntax for specifying neural network architectures inspired by that of popular deep learning libraries. A user need only specify a sequence of neural network layers and their respective hyperparameters and \textsc{ManimML\xspace{}} will automatically construct an animation of the entire network. Each layer is implemented separately, and each valid pair of layers has its own corresponding animation. This modular design allows for the easy addition of new layers and animations. 

\textbf{Usage Scenario.} Figure \ref{fig:teaser} demonstrates how Jane, an ML developer working with neural networks for character recognition, may use \textsc{ManimML\xspace{}} to communicate how her system works. Bob, a mobile app developer, is interested in incorporating Jane's character recognition system into his app. Bob is a technical person, but is largely unfamiliar with the inter-workings of machine learning systems. While Jane is an ML expert, she is much less familiar with advanced visualization libraries that she could use to develop a visual explanation, like D3. \textsc{ManimML\xspace{}} is ideal for this scenario. 

First, Jane can install \textsc{ManimML\xspace{}} using a simple one line command (\texttt{pip install manim\_ml}). 
Jane can then look at the specification of her model, which was implemented in a popular deep learning framework Pytorch and very simply copy the neural network architecture into a simple \textsc{ManimML\xspace{}} Python script (Figure \ref{fig:teaser}A). She decides that she wants to explain how the forward pass of her network works, so she writes a Python statement to do that (Figure \ref{fig:teaser}B). She can then run a simple command that render's the \textsc{ManimML\xspace{}} file (\texttt{manim -pqh network.py}). Finally, she can present the rendered video to Bob that visually explains the system she developed for Bob's mobile application. 

\section{Initial Feedback and Ongoing Work}

We believe that animation is a powerful medium for communicating scientific concepts. We are excited by the positive feedback we have gotten from the research community. At the time of writing, our GitHub repository has received over 1300 stars and amassed over 18,500 downloads on PyPi, the official third-party software repository for Python.\footnote{\url{https://pepy.tech/project/manim-ml}}  Further, demonstrations of our library have received hundreds of thousands of views on social media.\footnote{\url{https://twitter.com/alec_helbling/status/1609037745350926336}}
Some researchers have already used \textsc{ManimML\xspace{}} to generate visualizations for research papers.\footnote{\url{https://twitter.com/yanndubs/status/1597310609522950145}}

We plan to make two key extensions to this work. First, we want to horizontally expand the body of neural network architectures, layers, and animations that are available. For example, we want to include the ability to visualize decision trees and more advanced neural network architectures like transformers. Second, we would like to add systems for visualizing and interpreting specific instances of ML systems, rather than just architectures. For example, we plan on using a similar approach to BertViz \cite{vigMultiscaleVisualizationAttention2019} to visualize the self-attention layers of large language models. Finally, we wish to conduct extensive user studies to investigate \textsc{ManimML\xspace{}}'s efficacy and usability. We wish to observe how both beginner and seasoned ML experts use the tool and collect feedback on which additional functionality they wish to see \textsc{ManimML\xspace{}} support. 

\section{Acknowledgment}

This work was supported in part by DARPA GARD.

\vspace{-0.05in}

\bibliographystyle{abbrv-doi}

\bibliography{manual_bib}
\end{document}